\documentclass[10pt,twocolumn,letterpaper]{article}

\usepackage{cvpr}              

\definecolor{cvprblue}{rgb}{0.21,0.49,0.74}
\usepackage[pagebackref,breaklinks,colorlinks,allcolors=cvprblue]{hyperref}

\def\paperID{118} 
\def\confName{CVPR}
\def\confYear{2025}

\title{Towards Scale-Aware Low-Light Enhancement via Structure-Guided Transformer Design}

\author{Wei Dong$^*$ \quad Yan Min$^*$ \quad Han Zhou$^\dagger$ \quad Jun Chen \\
McMaster University \quad $^*$Equal Contribution \quad $^\dagger$Corresponding Author\\
{\tt\small \{dongw22, miny13, zhouh115, chenjun\}@mcmaster.ca}
}

\begin{document}

{
\twocolumn[{
\renewcommand\twocolumn[1][]{#1}
\maketitle
\vspace{-6mm}
\begin{center}

\setlength{\abovecaptionskip}{1.5mm}
\setlength{\parskip}{0mm} 
\setlength{\baselineskip}{0mm} 
\begin{minipage}[c]{1\textwidth}
    \includegraphics[width = 1\textwidth]{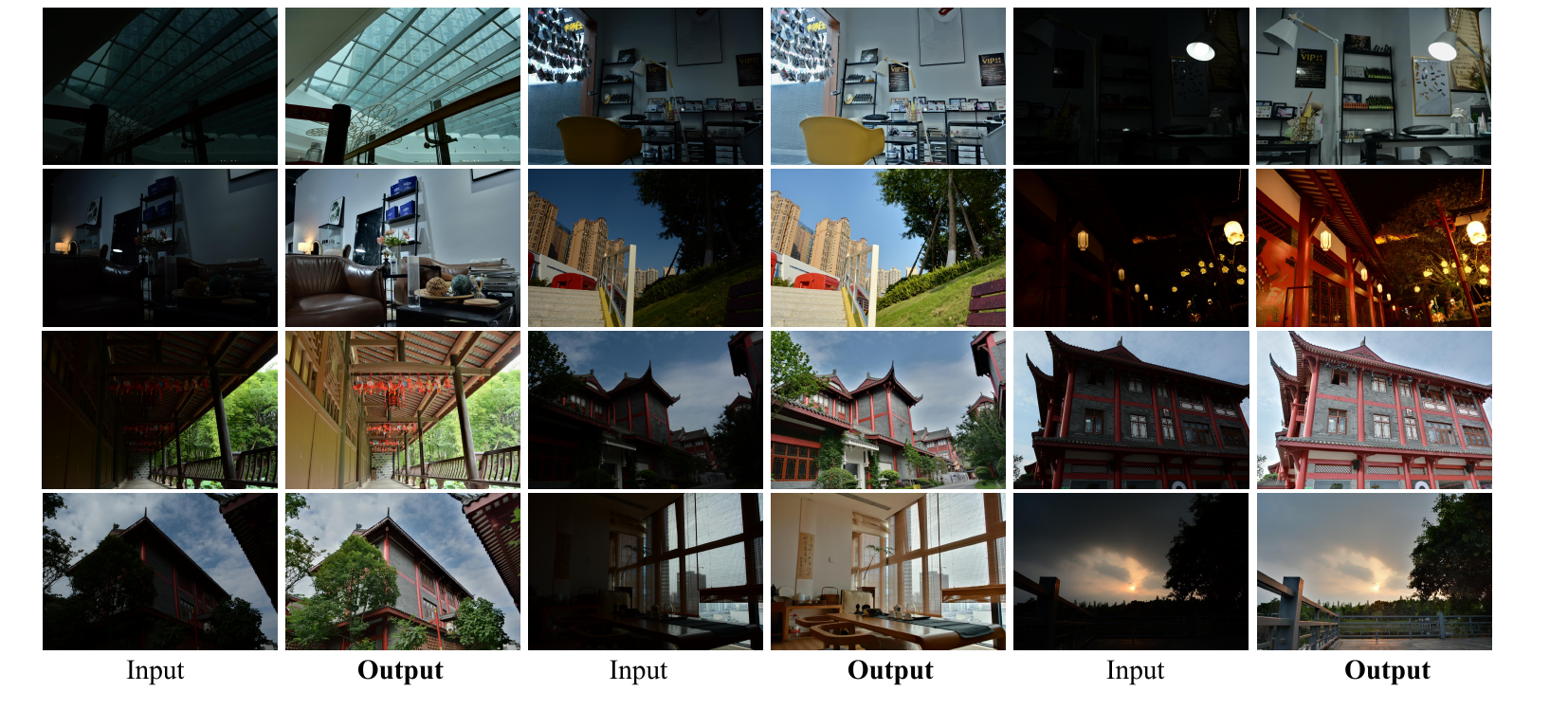}
    \setlength{\parskip}{0mm} 
\end{minipage}
\vspace{-5mm}
\captionof{figure}{Our enhanced results on \textit{NTIRE 2025 Low Light Enhancement Challenge}. Our method secures the \textbf{best PSNR}, achieves the \textit{second-best} overall performance, and effectively enhance low light inputs without over-exposed artifacts.}
\vspace{3mm}
\label{fig_first}
\end{center}}]
}

\begin{abstract}
Current Low-light Image Enhancement (LLIE) techniques predominantly rely on either direct Low-Light (LL) to Normal-Light (NL) mappings or guidance from semantic features or illumination maps. Nonetheless, the intrinsic ill-posedness of LLIE and the difficulty in retrieving robust semantics from heavily corrupted images hinder their effectiveness in extremely low-light environments. To tackle this challenge, we present \textbf{SG-LLIE}, a new multi-scale CNN-Transformer hybrid framework guided by structure priors. Different from employing pre-trained models for the extraction of semantics or illumination maps, we choose to extract robust structure priors based on illumination-invariant edge detectors. Moreover, we develop a CNN-Transformer Hybrid Structure-Guided Feature Extractor (HSGFE) module at each scale with in the UNet encoder-decoder architecture. Besides the CNN blocks which excels in multi-scale feature extraction and fusion, we introduce a Structure-Guided Transformer Block (SGTB) in each HSGFE that incorporates structural priors to modulate the enhancement process. Extensive experiments show that our method achieves state-of-the-art performance on several LLIE benchmarks in both quantitative metrics and visual quality. Our solution ranks second in the NTIRE 2025 Low-Light Enhancement Challenge. Code is released at \url{https://github.com/minyan8/imagine}.
\end{abstract}    
\section{Introduction}
\label{sec:intro}

Low Light Image Enhancement (LLIE) is an important task in computer vision. Normally captured images in dark environments often suffer from poor visibility, low contrast, and significant noise due to limited lighting conditions. These issues not only lead to the degradation of image qualities, but also hinder the performance of high-level vision tasks such as object segmentation, recognition, and tracking~\cite{segmentation, tracking, recognition}. Therefore, extensive research has been conducted to enhance low-light images and make them appear as if captured under better lighting conditions.

Most popular traditional methods for LLIE can be categorized into two types. One is histogram equalization-based methods ~\cite{abdullah2007dynamic, ibrahim2007brightness} , which adopt a straightforward strategy to perform gray-level remapping to enhance images with low visibility and contrast. However, these methods tends to introduce artifacts into the enhanced outputs. For example, in certain regions with uniform pixel values, the remapping process may produce excessively high or low pixel intensities, resulting in undesirably extremely bright or dark areas. Retinex-based approaches \cite{, fu2015probabilistic, wang2013naturalness, guo2016lime, park2017low, li2018structure} represent another popular class of traditional methods for LLIE. According to Retinex theory~\cite{land1977retinex}, a color image can be decomposed into two components: reflectance and illumination. Thus, the estimated illumination is essentially utilized as a physical prior to guide the enhancement process of the reflectance component. However, this strategy assumes that the reflectance is noise-free, which is not realistic in practical scenarios. Moreover, inaccurate illumination priors can lead to visual distortions and color inconsistencies.

\begin{figure}[t]
  \centering
  \setlength{\abovecaptionskip}{1.5mm}
  \includegraphics[width=0.85\linewidth]{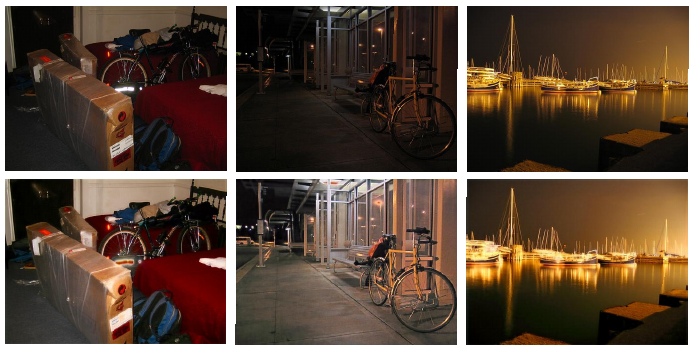}
  \caption{The enhancement results of Retinexformer~\cite{cai2023retinexformer} (pre-trained on LOL-v2-real dataset~\cite{lol-v2}) for real-world data. }
  \label{fig_reti_real}
  \vspace{-4mm}
\end{figure}

Recently, many deep learning models~\cite{hu2018exposure, gharbi2017deep, chen2018deep, deng2018aesthetic, yan2016automatic, xu2022snr, cai2023retinexformer, gpp25aaai} based on Convolutional Neural Network (CNN) and Transformers have been introduced for LLIE. However, end-to-end CNN methods often come with limited transparency and lacks solid theoretical backing, which can occasionally lead to unexpected or inconsistent results. While Transformer-based models excel at modeling non-local dependencies, directly applying Transformers~\cite{dosovitskiy2020image} to LLIE does not necessarily yield satisfactory results. Therefore, dedicated architectural designs are essential. To this end, Retinexformer~\cite{cai2023retinexformer} proposes to introduce the illumination prior into Transformers to guide the reflectance enhancement process and ultimately produce impressive well-lit images, which demonstrates the significant potential of employing physical priors in deep learning models.  However, we observe that the illumination prior utilized in Retinexformer is also learned from low light inputs via a lightweight neural network and there are no ground truth of illumination component to supervise such learning. Therefore, Retinexformer often struggles to generalize well in real-world scenarios and tends to produce unnatural color intensity and contrast, as presented in Fig.~\ref{fig_reti_real}, which motivates us to explore and integrate other more robust physical priors into deep learning models for LLIE.

In this paper, we propose \textbf{SG-LLIE}, a scale-aware CNN-Transformer framework for \textbf{LLIE} with the \textbf{G}uidance of \textbf{S}tructure priors. Overall, our method follows a UNet encoder–decoder architecture, and we develop the Hybrid Structure-Guided Feature Extractor (HSGFE) module at each level.  Within each HSGFE module, we first employ illumination-invariant edge detectors~\cite{geusebroek2001color} to extract stable structure priors, and then we introduce the Structure-Guided Transformer Block (SGTB) to incorporate structural priors into the restoration process. Besides, the CNN-based  Dilated Residual Dense Block (DRDB) and Semantic-Aligned Scale-Aware Module (SAM) proposed in~\cite{yu2022towards} are employed for multi-scale feature fusion. To summarize, we highlight three key contributions of this work:

\begin{itemize}
    \item We introduce \textbf{SG-LLIE}, a multi-scale CNN-Transformer hybrid framework guided by structure priors. 
    \item We first extract robust structure priors from low light images and then integrate these priors into our customized Transformer via structure-guided cross-attention, providing more effective guidance for LLIE.
    \item Our model shows solid performance in both quantitative metrics and visual results, taking the lead in the \textit{NTIRE 2025 Low-Light Image Enhancement Challenge} and staying competitive with other top solutions.
\end{itemize}

\section{Related Work}
\label{sec:related_work}

\begin{figure*}[t]
  \centering
  \includegraphics[width=1\textwidth]{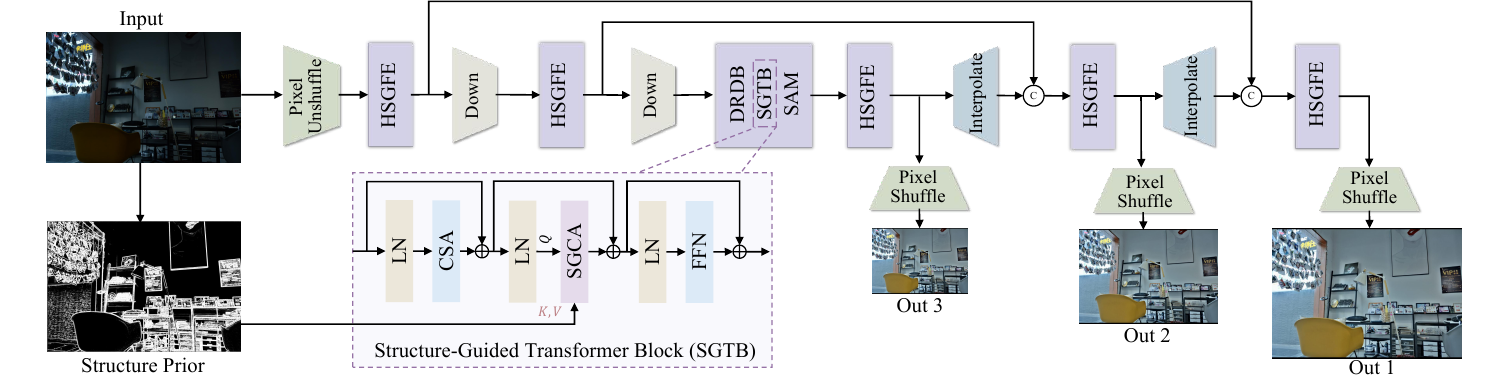}
  \caption{The overall framework of \textbf{SG-LLIE}. We develop our method based on ESDNet~\cite{yu2022towards} and adopt a similar UNet architecture. At each level of the encoder and decoder, our customized Hybrid Structure-Guided Feature Extractor (HSGFE) module is employed. Within each HSGFE, besides the Dilated Residual Dense Block (DRDB) and Semantic-Aligned Scale-Aware Module (SAM) proposed in ESDNet~\cite{yu2022towards}, we first extract structure priors based on color-invariant edge detectors~\cite{lengyel2021zero} and then develop the Structure-Guided Transformer Block (SGTB) to integrate these priors as guidance. With the integration of structure priors and our designed CNN-Transformer hybrid network, our method effectively improve the visibility and contrast with good noise suppression for diverse low light images.}
  \label{fig_frame}
\end{figure*}

\paragraph{Methods in LLIE} Histogram Equalization (HE)~\cite{abdullah2007dynamic, ibrahim2007brightness} and gamma correction~\cite{huang2012efficient, rahman2016adaptive} are widely used in traditional techniques. These methods perform well under relatively uniform lighting conditions. However, their effectiveness often degrades in real-world scenarios, where illumination in LL images is typically dynamic and diverse.  CNN-based models~\cite{hu2018exposure, gharbi2017deep, chen2018deep, deng2018aesthetic, yan2016automatic, zhou2023breaking, ntire23dehaze} has fundamentally altered the paradigm of image restoration and enhancement. LLNet~\cite{lore2017llnet} is the first to propose an autoencoder network, which enhances LL images without oversaturating the bright regions. Retinex theory are also widely applied in the deep learning-based methods. Yang \textit{et} al. \cite{yang2021sparse} introduces SGM-Net to combine priors and data-driven learning for LLIE, and effectively suppresses noise and improve contrast. However, due to the limitations of CNNs, these methods struggle to capture non-local information in images.

\paragraph{Transformers in Image Restoration}

Vision Transformers (ViTs) \cite{dosovitskiy2020image} are widely adopted in diverse low-level vision tasks \cite{cai2022degradation, chen2021pre, zamir2022restormer, hudson2021generative, dehazedct, shadowrefiner, ecmamba24neurips, ntire24dehaze, ntire24shadow, ntire2025day, ntire2025reflection, ntire2025shadow, rehit}. For example, Chen \textit{et} al. \cite{chen2021pre} introduces IPT, a pre-trained transformer model for low-level vision tasks like denoising, super-resolution, and deraining. Trained on corrupted image pairs from ImageNet, IPT can be fine-tuned for various tasks, not limited to image restoration. Conde \textit{et} al. \cite{conde2022swin2sr} proposes a network called Swin2SR, which builds upon SwinIR \cite{liang2021swinir} by incorporating Swin Transformer v2 \cite{swintransformer2} to enhance performance on compressed image super-resolution. In this work, we aim to combine CNN and Transformer to develop a hybrid network, which is capable of effectively capturing local and long range dependencies. 

\paragraph{Structure Prior} In recent years, image structure priors have been increasingly utilized in various low-level vision tasks, including image inpainting~\cite{dolhansky2018eye, nazeri2019edgeconnect, ren2019structureflow}, depth enhancement~\cite{gu2017learning, hui2016depth, li2016deep}, and image restoration~\cite{dogan2019exemplar, li2020enhanced, li2018learning, pan2014deblurring}. In a different line of work, Lengyel \textit{et al.} \cite{lengyel2021zero} proposes the integration of trainable, color-invariant edge detection layers into neural architectures to increase resilience to illumination variations. Alshammari \textit{et} al. \cite{alshammari2018impact} introduces the use of illumination-invariant image transforms to enhance scene understanding and segmentation in challenging lighting conditions. By combining invariant representations with chromatic cues, this approach improves the robustness of deep networks without altering their architecture, highlighting the value of pre-processing for handling illumination variation. In this work, we aim to extract robust structure priors from low light inputs and then integrate these priors to modulate the LLIE process.

\section{Method}
Our contribution mainly lies in the design of multi-scale CNN-Transformer hybrid network and the integration of illumination invariant structure priors. The framework of our method is illustrated in Figure~\ref{fig_frame}, where an encoder–decoder UNet architecture is designed. At each hierarchical level, the proposed Hybrid Structure-Guided Feature Extractor (HSGFE) functions as the core module, leveraging structural cues from the input image to facilitate the preservation of fine-grained details. In this section, We first discuss the structure prior extraction in Sec.~\ref{sec_struct}. Then, we introduce the details of our developed HSGFE module including the Structure-Guided Transformer Block (SGTB) and Structure-Guided Cross Attention (SGCA) in Sec.~\ref{sec_model}. Finally, the multi-scale loss functions are specified in Sec.~\ref{sec_loss}.

\subsection{Structure Prior Extraction}
\label{sec_struct}

To extract the structure prior, we adopt the Color Invariant Convolution (CIConv) proposed in~\cite{lengyel2021zero, lita25cvpr}, which serves as a task-adaptive edge detector. CIConv applies a learnable scale-aware transformation to the color-invariant representation of the input, producing a normalized edge response map that reflects task-relevant structure. Among the derived color-invariant representations based on the Kubelka-Munk (KM) reflection model~\cite{Kubelka-Munk, Kubelka-Munk-theory, Kubelka-Munk-theory2}, we adopt $\mathbf{W}$ as it provides robust edge detection under varying illumination, shading, and reflectance conditions:
\begin{equation}
    \mathbf{W}_{out} = \text{CIConv}(\mathbf{I}_{in}),
    \label{eq:0}
\end{equation}
where $\mathbf{I}_{in}$ denotes the input LL image. $\mathbf{W}_{out}$ represents the structural prior, which is later integrated into the Structure-Guided Transformer Block (SGTB) for guidance. The formula of CIConv can be expressed as:
\begin{equation}
\text{CIConv}(\mathbf{I}_{in}) = \frac{\log\left(\mathbf{W}^2(\mathbf{I}_{in}) + \epsilon\right) - \mu_{\mathcal{S}}}{\sigma_{\mathcal{S}}}
\end{equation}
where $\mu_{\mathcal{S}}$, $\sigma_{\mathcal{S}}$ and $\epsilon$ refer to the sample mean, standard deviation, and small perbutation. To compute $\mathbf{W}(\mathbf{I}_{in})$, we first use the Gaussian Color Model \cite{geusebroek2001color} to obtain the initial edge detectors, denoted as $\mathbf{E}$. Then, we use $\mathbf{E}$ to derive the second phase of edge detectors, denoted as $\mathbf{W}$, as follows:
\begin{align}
&\mathbf{W} = \sqrt{\mathbf{W}_x^2 + \mathbf{W}_{\lambda x}^2 + \mathbf{W}_{\lambda\lambda x}^2 + \mathbf{W}_y^2 + \mathbf{W}_{\lambda y}^2 + \mathbf{W}_{\lambda\lambda y}^2}, \\
&\mathbf{W}_x = \frac{\mathbf{E}_x}{\mathbf{E}}, \quad 
\mathbf{W}_{\lambda x} = \frac{\mathbf{E}_{\lambda x}}{\mathbf{E}}, \quad 
\mathbf{W}_{\lambda\lambda x} = \frac{\mathbf{E}_{\lambda\lambda x}}{\mathbf{E}}
\end{align}

\begin{figure}[t]
  \centering
  \setlength{\abovecaptionskip}{1.5mm}
  \includegraphics[width=1\linewidth]{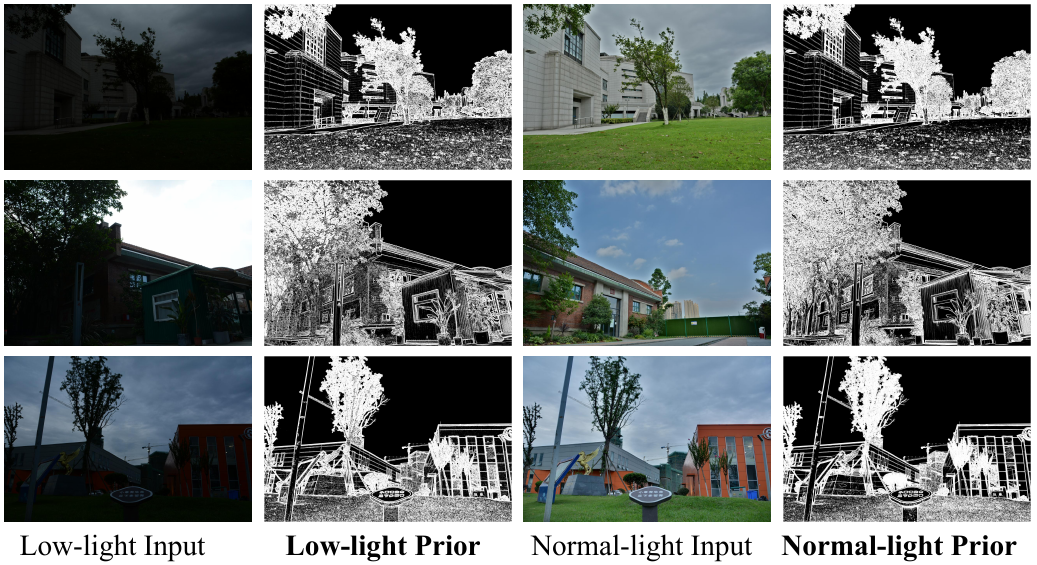}
  \caption{Example visualization of the extracted structure priors.}
  \label{fig_prior_compare}
\end{figure}

\paragraph{Physical Explanation} These equations mathematically indicates that $\mathbf{W}$ characterizes the spatial derivatives of spectral intensity. We visualize structure priors $\mathbf{W}$ for low-light and normal-light images in Fig.~\ref{fig_prior_compare}. It is clear that $W$ represents the stable edge and structure map across images with varying illumination conditions, highlighting its great potential to guide the enhancement process.

\begin{figure*}[t]
  \centering
  \setlength{\abovecaptionskip}{0.5mm}
  \includegraphics[width=1\linewidth]{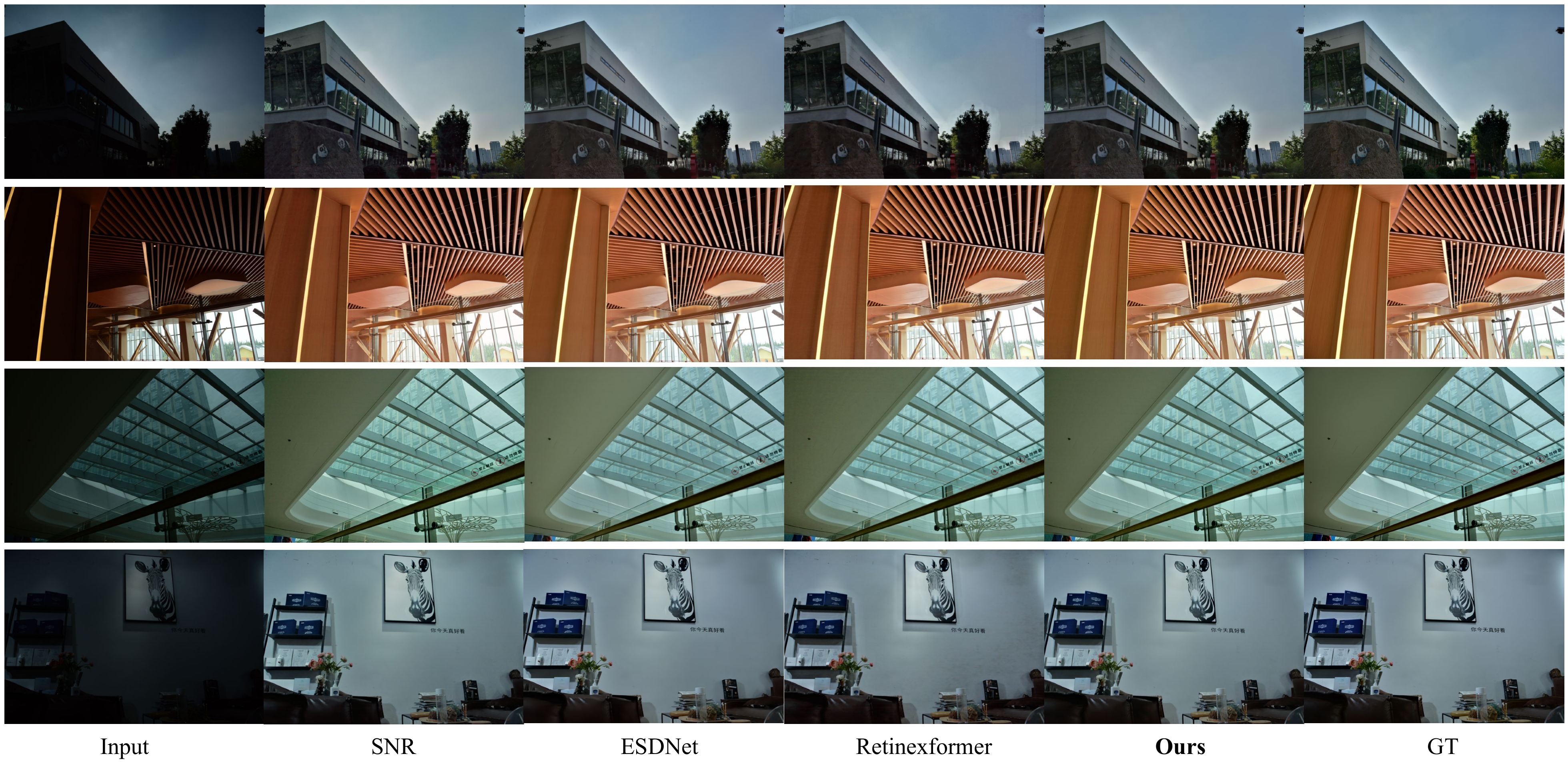}
  \caption{Qualitative comparisons on the \textit{NTIRE 2025 LLIE Challenge} dataset. We compare our model with SNR \cite{xu2022snr}, UHDM (SYSU) \cite{liu2024ntire}, and Retinexformer \cite{cai2023retinexformer}. Our model consistently performs best or comparably well. Please zoom in for a better view.}
  \label{fig_ntire25}
\end{figure*}

\subsection{Model Architecture}
\label{sec_model}

As illustrated in Fig.~\ref{fig_frame}, our model adopts the encoder-decoder architecture. The ``PixelUnShuffule'' and several convolutional layers (denoted as ``Down'' in Fig.~\ref{fig_frame}) are utilized for down-sampling, and the ``PixelShuffle'' or ``Interpolation'' process are used for up-sampling. Throughout the network, skip connections bridge corresponding encoder and decoder stages to maintain spatial coherence and support feature fusion. In each level in the encoding and decoding process, the Hybrid Structure-Guided Feature Extractor (HSGFE) module is proposed for representation learning and structure prior integration.

\paragraph{HSGFE Module} HSGFE Module is designed to exploit structural cues inherent extracted from the input for enhanced detail retention. The HSGFE begins by processing features through a Dilated Residual Dense Block (DRDB)~\cite{yu2022towards}, which enhances local representations. This is followed by a Structure-Guided Transformer Block (SGTB), where structural priors extracted in Sec.~\ref{sec_struct} are explicitly injected into the feature flow, guided by invariant edge descriptors \cite{geusebroek2001color}. Moreover, the Semantic-Aligned Scale-Aware Module (SAM) is incorporated to further accommodate scale diversity and adaptively fuse representations from different scales.  
Within SAM, residual connections are retained to maintain high-resolution spatial fidelity. 

\paragraph{Structure-Guided Transformer Block (SGTB)} As shown in Fig.~\ref{fig_frame}, our SGTB is composed of three main components: a Channel-wise Self-Attention (CSA), a Structure-Guided Cross Attention (SGCA), and a Feed-Forward Network (FFN). In addition, Layer Normalization is applied before each of these mechanisms, and three residual connections are applied to preserve residual information and support stable feature learning. Therefore, the feature flow for CSA can be represented as:
\begin{equation}
    \mathbf{F}_{out} = \text{CSA}(\text{LN}(\mathbf{F}_{in})) + \mathbf{F}_{in},
  \label{eq:e4}
\end{equation}
where \( \mathbf{F}_{\text{in}} \in \mathbb{R}^{H \times W \times C} \) and \( \mathbf{F}_{\text{out}} \in \mathbb{R}^{H \times W \times C} \) represent the input and output feature maps. We apply channel-wise attention here to let the model exchange information across channels, which facilitates capture long-range dependencies while ensuring model efficiency. Then, the processed representations are passed into the SGCA for further refinement.

\paragraph{Structure-Guided Cross Attention (SGCA)} This mechanism firstly reshape the input features \( \mathbf{F}_{\text{in}} \in \mathbb{R}^{H \times W \times C} \)  into \( \mathbf{X} \in \mathbb{R}^{HW \times C} \). Then, the resulting sequence \( \mathbf{X} \) is linearly projected to generate the Query (\( \mathbf{Q} \)) representation:
 \begin{equation}
    \mathbf{Q} = \mathbf{X} \mathbf{W}_{Q}^{\top}.
  \label{eq:e1}
\end{equation}

We observe that conventional low-light image enhancement models often distort the original structural details of input images. To address this issue, we incorporate structural priors into our attention mechanism to better preserve spatial consistency. Specifically, structural priors are introduced into the Structure-Guided Cross Attention (SGCA) block to provide the \textbf{K}ey and \textbf{V}alue elements, denoted as \( \mathbf{K}_{p} \) and \( \mathbf{V}_{p} \), where the subscript \( p \) refers to structure \textbf{p}riors. These two representations are obtained by:
 \begin{equation}
    \mathbf{K_p} = \mathbf{X} \mathbf{W}_{K_p}^{\top}, \quad
    \mathbf{V_p} = \mathbf{X} \mathbf{W}_{V_p}^{\top},
  \label{eq:e2}
\end{equation}
where $\mathbf{W}_{K_p}\in \mathbb{R}^{C \times C}$ and $\mathbf{W}_{V_p}\in \mathbb{R}^{C \times C}$ are learnable parameter matrice. We then formulate our structure-guided attention mechanism as follows:
\begin{equation}
    \text{Attention}(\mathbf{Q}, \mathbf{K}_p, \mathbf{V}_p) = \text{softmax}(\frac{\mathbf{Q} \cdot \mathbf{K}_p}{\lambda}) \cdot \mathbf{V}_p,
  \label{eq:e3}
\end{equation}
where $\lambda$ is a learnable parameter that adaptively adjusts the scale of the matrix multiplication. To this end, we design our cross-attention mechanism to not only gather long-range dependencies but also blend structural cues directly into the current feature representations.

\begin{table*}[t]
\centering
\renewcommand{\arraystretch}{1.0}
\scriptsize
\resizebox{0.6\textwidth}{!}{
\begin{tabular}{l|c|c|c|c|c}
\hline
\textbf{Team} & \textbf{Rank} & \textbf{PSNR↑} & \textbf{SSIM↑} & \textbf{LPIPS↓} & \textbf{NIQE↓} \\
\hline
NWPU-HVI & 1 & 26.24 & 0.861 & 0.128 & 10.9539 \\
\textbf{Imagine(Ours)} & \textcolor{red}{2} & \textcolor{red}{26.345} & 0.858 & 0.133 & 11.8073 \\
pengpeng-yu & 3 & 25.849 & 0.858 & 0.134 & 11.2933 \\
DAVIS-K & 4 & 25.138 & 0.863 & 0.127 & 10.5814 \\
SoloMan & 5 & 25.801 & 0.856 & 0.130 & 11.4979 \\
Smartdsp & 6 & 25.47 & 0.848 & 0.120 & 10.5387 \\
Smart210 & 7 & 26.148 & 0.855 & 0.137 & 11.5165 \\
WHU-MVP & 8 & 25.755 & 0.855 & 0.138 & 11.2140 \\
BUPTMM & 9 & 25.673 & 0.855 & 0.137 & 11.2831 \\
NJUPTIPR & 10 & 25.011 & 0.848 & 0.122 & 10.1485 \\
SYSU-FVL-T2 & 11 & 25.652 & 0.857 & 0.135 & 11.5897 \\
\hline
\end{tabular}
}
\caption{Performance comparison of the top 11 teams in the \textit{NTIRE 2025 LLIE Challenge}. Metrics include PSNR, SSIM, LPIPS (lower is better), and NIQE (lower is better). Our team (\textbf{Imagine}) achieved the best \textit{PSNR} and the second-best overall score among 28 teams.}
\label{tab:ntire_llie_results}
\end{table*}

\definecolor{best}{rgb}{1,0,0}
\begin{table*}[ht]
\centering
\small
\setlength{\tabcolsep}{3pt}
\renewcommand{\arraystretch}{1.1}
\begin{tabular}{l|ccc|ccc|ccc|cc}
\toprule
\textbf{Methods} & \multicolumn{3}{c|}{\textbf{LOL-v1}} & \multicolumn{3}{c|}{\textbf{LOL-v2-real}} & \multicolumn{3}{c}{\textbf{NTIRE 2025 LLIE}} &\multicolumn{2}{c}{\textbf{Complexity}} \\
\hline
& PSNR$\uparrow$ & SSIM$\uparrow$ & LPIPS$\downarrow$ & PSNR$\uparrow$ & SSIM$\uparrow$ & LPIPS$\downarrow$ & PSNR$\uparrow$ & SSIM$\uparrow$ & LPIPS$\downarrow$ & Params (M) & FLOPs \\
\midrule

KinD \cite{zhang2019kindling} & 20.87 & 0.802 & 0.207 & 17.54 & 0.669 & 0.365 & 21.29 & 0.829 & 0.186 &8.02 & 34.09 G \\
MIRNet \cite{Zamir2020MIRNet} & 24.14 & 0.832 & 0.131 & 19.35 & 0.708 & 0.138 & 24.21 & 0.857 & 0.141 &31.76 & 785 G\\
SNR \cite{xu2022snr} & 24.61 & 0.842 & 0.151 & 20.82 & 0.811 & 0.161 & 24.39 & 0.878 & 0.149 &4.01 & 26.35 G \\
Restormer \cite{Zamir2021Restormer} & 22.43 & 0.823 & 0.147 & 18.69 & 0.834 & 0.232 & 22.47 & 0.846 & 0.145 &26.13 & 144.25 G\\
Retinexformer \cite{cai2023retinexformer} & \textcolor{best}{25.16} & 0.845 & 0.131 & \textcolor{blue}{22.79} & 0.840 & 0.171 & 25.06 & 0.872 & 0.183 &1.61 & 17.02 G\\
LLFlow \cite{wang2021low} & 21.09 & 0.861 & 0.116 & 17.43 & 0.831 & 0.129 & 23.49 & 0.880 & 0.131 &17.42 &1.05 T\\
ESDNet \cite{yu2022towards} & 23.10 & 0.846 & 0.094 & 20.48 & 0.841 & 0.134 & 26.26 & 0.892 & \textcolor{best}{0.111} &10.62 &28.82 G\\
GLARE \cite{glare24eccv} & 23.55 & \textcolor{blue}{0.863} & \textcolor{best}{0.086} & 22.75 & \textcolor{blue}{0.856} & \textcolor{best}{0.105} & 25.35  & 0.887 & 0.122  &24.74 &223 G\\

\hline
\textbf{Ours} & \textcolor{blue}{24.63} & \textcolor{best}{0.873} & \textcolor{blue}{0.092} & \textcolor{best}{22.84} & \textcolor{best}{0.859} & \textcolor{blue}{0.126} & \textcolor{best}{26.75} & \textcolor{best}{0.899} & \textcolor{blue}{0.113} &12.67 &35.80 G\\
\bottomrule
\end{tabular}
\caption{Quantitative comparison on LOL-v1, LOL-v2-real, and \textit{NTIRE 2025} datasets. [Key: \textcolor{red}{Best Performance}; \textcolor{blue}{Second-Best Performance}]}
\label{tab_main}
\end{table*}

\paragraph{Scale-Adaptive Neural Architecture} Inspired by the approach in Yu \textit{et} al. \cite{yu2022towards}, we integrate a Semantic-Aligned Scale-Aware Module (SAM) following the SGTB within our HSGFE module to effectively extract features across multiple scales. In real-world scenarios, images are often captured at varying resolutions (\textit{e.g.}, 6000×4000 or 2992×2000), which poses challenges for consistent feature representation. To address this, SAM leverages a combination of pyramid-based feature extraction and cross-scale dynamic fusion.

Initially, the input feature map $\mathbf{F}_{in,0} \in \mathbb{R}^{H \times W \times C}$ undergoes bilinear interpolation to produce two additional versions at coarser scales: $\mathbf{F}_{in,1} \in \mathbb{R}^{\frac{H}{2} \times \frac{W}{2} \times C}$ and $\mathbf{F}_{in,2} \in \mathbb{R}^{\frac{H}{4} \times \frac{W}{4} \times C}$. These multi-resolution feature maps are then processed independently through convolutional layers to yield the corresponding pyramid representations: $\mathbf{Y}_{in,0}$, $\mathbf{Y}_{in,1}$, and $\mathbf{Y}_{in,2}$.

Subsequently, cross-scale fusion is performed to integrate the multi-resolution feature maps. To achieve this, each scale-specific feature map is assigned a learnable weight matrix $\alpha_i$, where $i = 0, 1, 2$. These weights are derived by applying global average pooling to each of the three feature maps independently. The resulting pooled features are then passed through a multi-layer perceptron (MLP), which facilitates adaptive and effective cross-scale interaction. This process can be formulated as follows: 
\begin{equation}
    [\beta_0, \beta_1, \beta_2 = \text{MLP}(\alpha_{0},\alpha_{1},\alpha_{2})].
  \label{eq:e5}
\end{equation}

Finally, the fused feature map $\mathbf{F}_f$ is obtained as follows, and $\mathbf{F}_{in,0}$ is added to retain more input information.
\begin{equation}
    \mathbf{F}_f = \mathbf{F}_{in,0}+\beta_0\odot \mathbf{Y}_{in,0} +\beta_1\odot \mathbf{Y}_{in,1}+\beta_2\odot \mathbf{Y}_{in,2}.
  \label{eq:e6}
\end{equation}

\subsection{Loss Function and Adaptive Adjustment Layer}
\label{sec_loss}

\paragraph{Loss Function} As Fig.~\ref{fig_frame} shows, for network training, we design the loss function based on three levels of output images—$\hat{\mathbf{I}}_{1}$, $\hat{\mathbf{I}}_{2}$, and $\hat{\mathbf{I}}_{3}$—which correspond to the output images at three different resolutions from the respective levels of our decoder:
\begin{align}
    L_{\text{total}} = & \sum_{i=1}^{3} L_C\left(\mathbf{I}_i, \hat{\mathbf{I}}_i\right) 
    + \lambda \cdot \sum_{i=1}^{3} L_P\left(\mathbf{I}_i, \hat{\mathbf{I}}_i\right) \notag \\
    & + \gamma \cdot \sum_{i=1}^{3} L_{\text{MS-SSIM}}\left(\mathbf{I}_i, \hat{\mathbf{I}}_i\right),
    \label{eq:e7}
\end{align}
where ${L}_{1}$, ${L}_{\text{P}}$, and ${L}_{\text{MS-SSIM}}$ represents the Charbonnier loss \cite{lai2018fast}, perceptual loss \cite{johnson2016perceptual}, and Multi-Scale SSIM (MS-SSIM) loss, respectively. The weighting factors are set as $\lambda = 0.01$ and $\gamma = 0.4$. 

\paragraph{Adaptive Adjustment Layer} When evaluating our method on the dataset provided by NTIRE 2025 Low-Light Enhancement Challenge, we observe the illumination of enhanced results differs a lot even for the training data. We attribute this to the diverse light condition within the training set and we aim to develop an additional layer for slight illumination adjustment. Specifically, by analyzing the illumination level of the training set (including the input and enhanced results of our proposed SG-LLIE), we classify all training samples into five classes, and employ a specific adjustment layer between 0.96 and 1.05 to each class. Moreover, three linear layers connected by ReLU is employed to learn this classification pattern and the cross-entropy is utilized for the optimization of this adaptive adjustment layer. Finally, we add this adjustment layer into SG-LLIE and optimize SG-LLIE using a low learning rate ($2\times 10^-7$) without updating the adaptive adjustment layer. Besides, only $\hat{\mathbf{I}}_{1}$ is leveraged as the loss function in this stage.

\section{Experiments}

\subsection{Performance on NTIRE 2025 LLIE Challenge}

\begin{figure*}[t]
  \centering
  \setlength{\abovecaptionskip}{1.5mm}
  \includegraphics[width=1\linewidth]{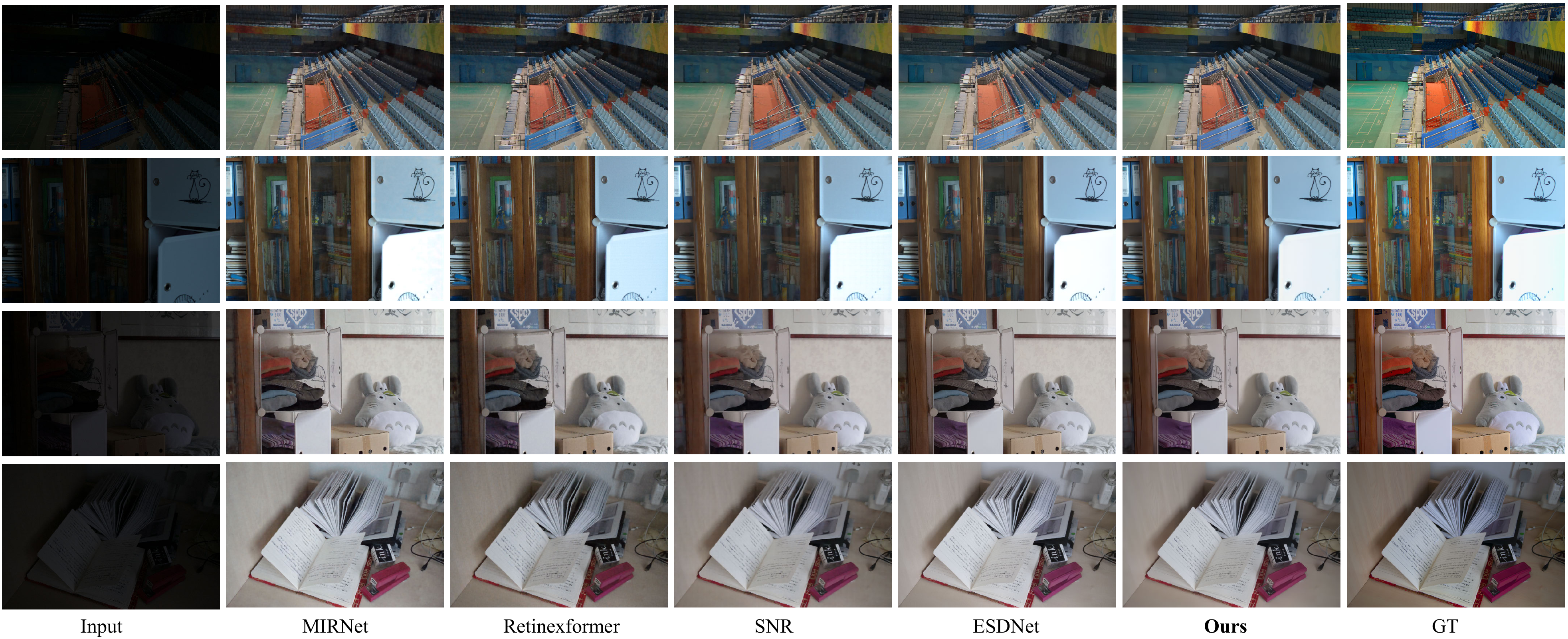}
  \caption{Visual comparisons on the LOL-v1 dataset, with MIRNet \cite{Zamir2020MIRNet}, RetinexFormer \cite{cai2023retinexformer}, SNR \cite{xu2022snr}, and ESDNet~\cite{yu2022towards}.}
  \label{fig_lol-v1}
  \vspace{-2mm}
\end{figure*}

\paragraph{Dataset} We use the dataset provided by the \textit{NTIRE 2025 Low Light Enhancement Challenge} \cite{liu2025ntire}. This dataset consists of 219 training images, 46 validation images, and 30 test images. Most images have a resolution of 2992×2000, with several reaching up to 6000×4000. Compared to other low-light image datasets, such as LOL \cite{lol} and LOL-v2-real \cite{lol-v2}, the \textit{NTIRE} dataset offers higher-resolution images and more diverse content, better reflecting real-world captures from modern smartphones. Both indoor and outdoor scenes are included. Besides, we also use the training set from the \textit{NTIRE 2024 Low-Light Enhancement Challenge} \cite{liu2024ntire} for fine-tuning.

\paragraph{Implementation Details} We develop and train our model using PyTorch on a single A100 GPU, without any pretrained weights. The model is optimized with the Adam optimizer \cite{kingma2014adam} $(lr=2\times10^{-4},\beta_1=0.9,\beta_2=0.999)$, and gradient clipping is applied. Training runs for 15{,}000 iterations with a fixed patch size of 1600\(\times\)1600 and a batch size of 1. The initial learning rate is set
as $2\times10^{-4}$ and the cyclic cosine annealing schedule~\cite{loshchilov2016sgdr} is employed. Data augmentation includes geometric transforms and mixup with \(\beta=1.2\) and identity mapping enabled. The training of the adjustment layer and the fine-tuning of SG-LLIE is provided in Sec.~\ref{sec_loss}.

\paragraph{Quantitative and Qualitative on NTIRE 2025 LLIE Challenge} As presented in Tab.~\ref{tab:ntire_llie_results}, our method achieves the best PNSR, the third best SSIM in \textit{NTIRE 2025 LLIE Challenge}. Overall, our method ranks 2nd out of 28 teams from around the world, highlighting the outstanding performance of our proposed approach. The enhancement outputs of our method are presented in Fig.~\ref{fig_first}, which demonstrates that our method can handle diverse low light images captured with varying illuminations and indoor or outdoor scenarios. Both quantitative and qualitative results demonstrate the effectiveness of our proposed method.

\subsection{Performance on LLIE Benchmark Datasets} 

\paragraph{Experiment Settings} To comprehensively evaluate the performance of our method, we compare our method with KinD \cite{zhang2019kindling}, MIRNet \cite{Zamir2020MIRNet}, Restormer \cite{zamir2022restormer}, LLFlow \cite{wang2021low}, Retinexformer \cite{cai2023retinexformer}, SNR \cite{xu2022snr}, ESDNet (refined by Team SYSU-FVL-T2~\cite{liu2024ntire} in \textit{NTIRE 2024 LLIE Challenge}), and GLARE~\cite{glare24eccv} on the \textit{NTIRE 2025 LLIE} dataset. Due to the unavailability of the ground truth for validation and test set, we build a test set includes 11 images with scenes that are similar but not identical to those in the training set. We construct this test set by analyzing the differences between the \textit{NTIRE 2024 LLIE} and \textit{NTIRE 2025 LLIE} training sets. For a fair comparison, we use a patch size of $512 \times 512$ and a batch size of 2 during training on the NTIRE 2025 LLIE dataset for all methods being compared. Besides the NTIRE dataset, we also evaluate our method on two additional datasets: LOL-v1 \cite{lol} and LOL-v2-real \cite{yang2021sparse}, where the patch size and batch size are set to $384 \times 384$ and 4, respectively. Notably, the adjustment layer is not included in this setting for a fair comparison.

\begin{figure*}[t]
  \centering
  \setlength{\abovecaptionskip}{2mm}
  \includegraphics[width=1\linewidth]{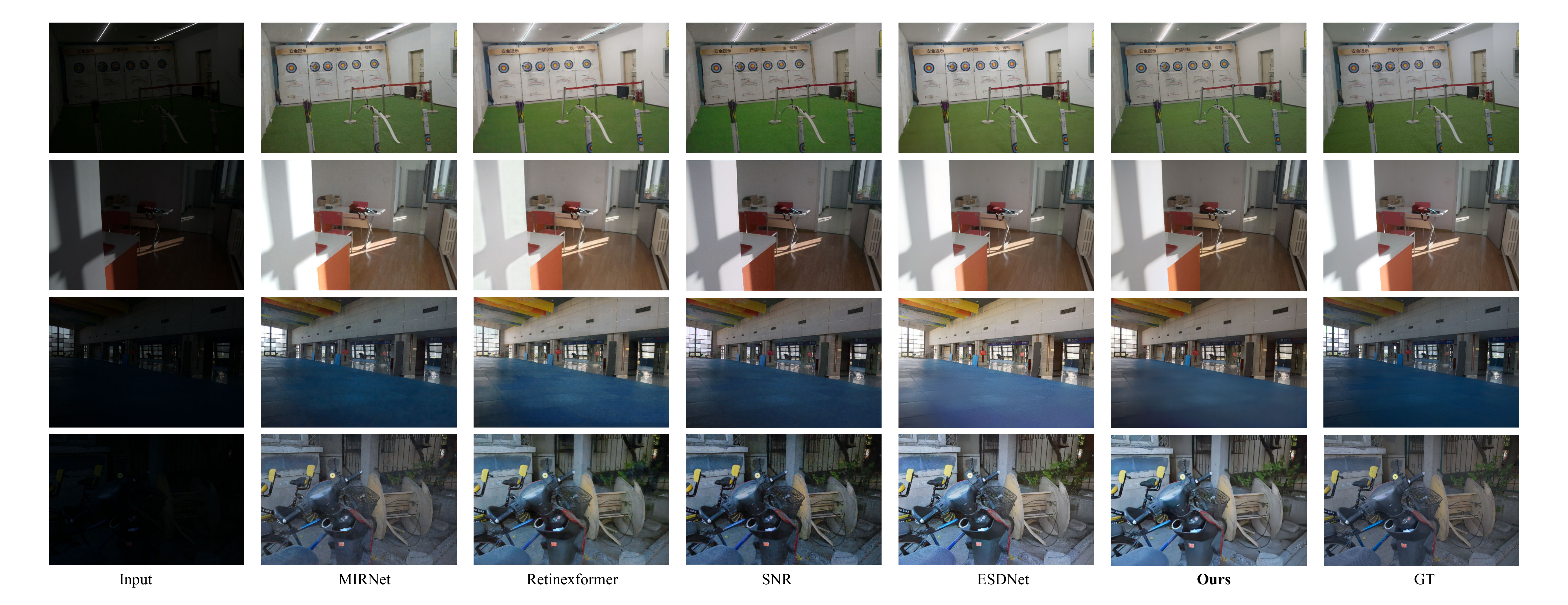}
  \caption{Visual comparisons on LOL-v2-real dataset with MIRNet \cite{Zamir2020MIRNet}, RetinexFormer \cite{cai2023retinexformer}, SNR \cite{xu2022snr}, and ESDNet~\cite{yu2022towards}.}
  \label{fig_lol-v2real}

\end{figure*}

\paragraph{Quantitative Results} The quantitative comparisons on the \textit{NTIRE 2025 LLIE}, LOL-v1, and LOL-v2-real datasets are shown in Tab~\ref{tab_main}. On LOL-v1, we obtain the highest SSIM (0.873) and achieve the second-bset performance on PSNR and LPIPS, indicating superior structural fidelity and perceptual quality, while maintaining a competitive PSNR (24.63). For LOL-v2-real, our approach continues to lead with a PSNR of 22.84, SSIM of 0.859, and LPIPS of 0.126, surpassing all other methods. On the \textit{NTIRE 2025 LLIE} benchmark, our method again outperforms all baselines with the highest PSNR (26.75), SSIM (0.899), and second-best LPIPS (0.113), closely following ESDNet (0.111). Although Retinexformer~\cite{cai2023retinexformer} achieves higher PSNR scores with fewer parameters on the LOL-v1 and LOL-v2-real datasets, it falls short in recovering structural details compared to our method. In addition, its performance on high-resolution images degrades noticeably, and it incurs higher GPU memory consumption owing to its fully Transformer-based design. In contrast, our method outperforms both GLARE~\cite{glare24eccv} and LLFlow~\cite{wang2021low} in terms of computational efficiency and enhancement performance. Furthermore, our SG-LLIE introduces only approximately 2M additional parameters compared to ESDNet~\cite{yu2022towards}, yet achieves an average gain of 1.46 dB in PSNR and 0.017 in SSIM.

\paragraph{Qualitative Results}

We present visual comparisons of our model with the methods listed in Table~\ref{tab_main}. As shown in Figure~\ref{fig_lol-v1}, Figure~\ref{fig_lol-v2real}, and Figure~\ref{fig_ntire25}, our results are perceptually better or comparable to those of other methods. Our method more effectively captures underlying lighting conditions and preserves the original content of the input images. While MIRNet~\cite{Zamir2020MIRNet} and SNR \cite{xu2022snr} occasionally produce slightly brighter outputs, our model delivers a more perceptually pleasing balance of contrast and natural appearance, particularly in terms of retaining underlying image information.

\subsection{Ablation Study}
\label{sec_ex_abla}

\begin{table}[t]
\centering
\renewcommand\arraystretch{1.2}
\setlength{\tabcolsep}{3pt}
\begin{tabular}{c ccc}
\toprule
\textbf{Configuration} & PSNR$\uparrow$ & SSIM$\uparrow$ & LPIPS$\downarrow$ \\
\midrule
w/o SGCA  & 23.67 & 0.855 & 0.095 \\
w/o SGTB & 23.38 & 0.849 & 0.096 \\
w/o MS-SSIM & 24.48 & 0.868& 0.093 \\
Full Model & \textbf{24.63} & \textbf{0.873} & \textbf{0.092}\\
\bottomrule
\end{tabular}
\caption{Ablation results on LOL-v1 dataset. Our full model achieves the best performance in terms of all metrics and removing any component our complete model leads to obvious performance drop, highlighting the rationality of the design of our method.}
\vspace{0mm}

\label{tab_abla}
\end{table}

To verify the contribution of each component in our method, we conduct ablation studies on the LOL-v1 dataset \cite{chen2018deep}.

\paragraph {Structure-Guided Cross Attention (SGCA)} To study the importance of the extracted structure prior and our proposed SGCA, we remove these two parts from our method. Tab.~\ref{tab_abla} reports the quantitative performance of this modification on LOL-v1 dataset, which still achieve competitive enhancement performance on all measured metrics compared to ESDNet, Restormer, and MIRNet in Tab.~\ref{tab_main}. However, compared to this modified version, our full model shows significantly enhanced SSIM and PSNR scores by integrating structure prior using our proposed SGCA. This comparisons manifest the importance of our proposed SGCA and the structure prior generated by the extraction pipeline discussed in Sec.~\ref{sec_struct}.

\paragraph {Structure-Guided Transformer Block (SGTB)} Similarly, we implement an adaptation to illustrate the effectiveness of the developed SGTB. Specifically, we remove the SGTB from our customized Hybrid Structure-Guided Feature Extractor (HSGFE) module. The quantitative results of the remaining network are reported in Tab.~\ref{tab_abla}. The discernible performance gap between this configuration and the full model (\textit{i.e.}, a 1.25 dB drop in PSNR and a 0.024 drop in SSIM) demonstrates the superiority of our proposed Structure-Guided Transformer Block (SGTB). It is worth noting that the remaining network is entirely CNN-based, and its relatively poor performance underscores the importance of our hybrid CNN-Transformer architecture.

\paragraph {Loss Function} As introduced in Sec.~\ref{sec_loss}, we add the muti-scale MS-SSIM loss into our complete optimization objective, which is not included in ESDNet. To verify the effectiveness of this new introduced loss function, we remove it and adopt the same optimization strategy in ESDNet. The quantitative results are reported in Tab.~\ref{tab_abla}, which demonstrate that the integration of multi-scale MS-SSIM loss helps achieve higher SSIM performance.

\section{Conclusion}

We present SG-LLIE, a CNN-Transformer hybrid encoder-decoder architecture augmented with Hybrid Structure-Guided Feature Extractor (HSGFE) modules to effectively enhance low-light images. These modules leverage structural cues, extracted from the input by color-invariant edge detector, to preserve fine-grained details, combining dilated residual dense blocks with transformer layers guided by structural priors. To further address scale variation, the scale-aware semantic-aligned module is incorporated, and skip connections are employed to preserve spatial consistency throughout the network. Our design enables robust and detail-aware low-light enhancement across diverse lighting conditions. Extensive experiments validate the effectiveness of our method, achieving the \textbf{best PSNR score} and \textbf{second-best overall performance} among 28 teams in \textit{NTIRE 2025 Low Light Enhancement Challenge} \cite{liu2025ntire}.
\clearpage

\clearpage
{
    \small
    \bibliographystyle{ieeenat_fullname}
    \bibliography{main}
}

\end{document}